\documentclass[11pt]{article}
\usepackage[preprint]{acl}
\usepackage{tikz}
\usepackage{multirow}

\usetikzlibrary{arrows.meta, positioning}

\usepackage{times}
\usepackage{latexsym}
\usepackage{amsmath} 
\usepackage{booktabs}

\usepackage[T1]{fontenc}

\usepackage[utf8]{inputenc}

\usepackage{microtype}

\usepackage{inconsolata}

\usepackage{graphicx}

\title{GRPO for Financial Advice Generation: Outperforming Commercial LLMs under CATE Evaluation}

\author{%
  \textbf{Ofir Ben Shoham, Shrutendra Harsola, Vignesh Subrahmaniam,} \\
  \textbf{Shravan Mohan, Yakov Gazman, Oded Vainas} \\
  Intuit \\
  \texttt{\{ofir\_benshoham, shrutendra\_harsola, vignesh\_subrahmaniam,} \\
  \texttt{shravan\_mohan, yakov\_gazman, oded\_vainas\}@intuit.com}
}

\begin{document}
\maketitle
\begin{abstract}
  Generating actionable financial advice from business records demands that models integrate numerical reasoning, domain knowledge, and sound judgment, while avoiding recommendations that could harm the business. Direct supervision is difficult: historical decisions are not necessarily optimal, and high-quality free-form labels are expensive to obtain. We formulate financial advice generation as a reinforcement learning problem and fine-tune an open-weight language model using Group Relative Policy Optimization (GRPO). Our reward is an LLM-as-a-judge rubric that scores each recommendation across multiple binary dimensions of advice quality, augmented with a safety gate for harm prevention.

  Since LLM-based evaluation alone cannot confirm whether improvements reflect genuine business value rather than adaptation to the judge, we complement it with a judge-independent audit based on a standard doubly-robust Conditional Average Treatment Effect (CATE) estimator. Under this observational off-policy audit, our trained LLM achieves approximately twice the estimated gross-profit lift of the strongest evaluated commercial baseline ($0.0228$ vs.\ $0.0104$), together with the lowest downside rate and the least negative tail risk of any policy evaluated. Notably, the two evaluations do not rank the baselines identically: the untrained base model places last on the judge rubric but second on the causal audit, indicating that the audit captures a signal the judge does not. Our results demonstrate that GRPO with a finance-grounded reward signal can produce substantially more useful business recommendations than commercial LLMs, and that a judge-independent causal audit is a valuable complement to, rather than a confirmation of, LLM-as-a-judge assessment in financial NLP.
\end{abstract}

\section{Introduction}
Businesses generate large volumes of financial records, yet translating those records into concrete, actionable advice remains a difficult problem. A useful recommendation must be grounded in the specific numbers of the business, aligned with its financial goals, realistically actionable, and safe to implement. These requirements make financial advice generation challenging: outputs are open-ended, quality is multidimensional, and a bad recommendation carries real business risk.

Supervised approaches face a fundamental obstacle: historical business decisions are not optimal gold labels. In addition, collecting high-quality free-form annotations from domain experts is expensive and does not scale. In contrast, Reinforcement Learning offers a natural alternative: instead of imitating past decisions, a model can be trained directly to maximize a reward that captures the properties we care about \cite{kaelbling1996reinforcement}.

Therefore, we formulate financial advice generation as an RL problem and fine-tune an open-weight language model using Group Relative Policy Optimization (GRPO) \citep{shao2024deepseekmath}. Following the RLAIF paradigm \citep{lee2023rlaif}, our reward is provided by an LLM-as-a-judge rubric that scores each recommendation across multiple binary dimensions of advice quality, augmented with a safety gate for harm prevention.

A key concern with judge-based training is reward hacking: a model may learn to satisfy the judge without producing recommendations that are genuinely useful. To address this, we complement judge-based evaluation with a judge-independent audit based on a standard doubly-robust Conditional Average Treatment Effect (CATE) estimator \cite{abrevaya2015estimating}, which estimates the expected impact of following a recommendation on gross profit from observational data. Under this audit, our GRPO-trained model achieves approximately twice the gross-profit lift of the strongest evaluated commercial baseline, with the lowest downside rate and least negative tail risk of any policy evaluated. The audit further reveals that judge scores and estimated business value are only loosely coupled across the baselines, which supports its use as a genuinely independent check rather than a corroboration of the judge.

In this work, our main contributions are: (1) \textbf{rubric-grounded, safety-gated GRPO for financial advice}: we formulate free-form business financial advice generation as an RL problem and fine-tune an open-weight LLM with a finance-specific, safety-gated rubric reward; (2) a \textbf{natural-language-to-action causal audit}: we map generated advice into a fixed business-action taxonomy and apply a standard doubly-robust CATE estimator to audit whether recommendations correspond to historically beneficial actions; and (3) a \textbf{dual evaluation against commercial and open-weight baselines}, using both judge-based rubric scores and a judge-independent score from a standard doubly-robust CATE estimator, with uncertainty, downside-rate, and tail-risk metrics; we show that the two evaluations agree on our policy but diverge on the baselines, and argue that this divergence is itself evidence that the audit is not a proxy for the judge.

\section{Related Work}
Reinforcement learning from human or AI feedback has been widely used to align LLM outputs with desired properties \citep{lee2023rlaif}. GRPO \citep{shao2024deepseekmath} avoids the need for a separate value model by computing relative advantages within a group of sampled outputs. While GRPO was originally introduced for mathematical reasoning, it has since been applied to broader domains using structured rubric rewards. Most closely to our setting, \citet{bhattarai2026rubric} use multi-criterion rubric rewards with GRPO for scientific reasoning. In the financial domain, RL has been applied to tasks such as alpha factor screening \citep{jiang2025alpha} and trading, but open-ended business advice generation has not been addressed.

Financial NLP work has largely focused on benchmarking general-purpose LLMs, finding that they still struggle with money-related reasoning \citep{rosero2025evaluating, klimaszewski2025avenibench} and that traditional models can outperform generative LLMs when numeric signals are central \citep{drinkall2025forecasting}. These results highlight the difficulty of the financial domain, which motivates training a model specifically for business advice, both in terms of advice quality and in terms of the latency and cost overhead of relying on general-purpose commercial LLMs.

To train such a model without gold labels, we rely on an LLM judge as a reward signal. However, improvements driven by a judge reward may reflect adaptation to the judge's preferences rather than genuine business value, and LLM-based evaluators are known to overestimate systems that align with their style \citep{lee2023rlaif}. We therefore complement judge-based evaluation with a judge-independent audit based on a standard doubly-robust CATE estimator \citep{robins1994estimation, bang2005doubly, dudik2011doubly, abrevaya2015estimating}.

\section{Method}

\subsection{Problem Formulation}
\label{sec:formulation}

We are given a business financial state $s$, summarizing a business's recent financials (revenue, cost of goods sold, operating expenses, vendor and product line items, and recent trends), together with a target goal $g$ (we focus on gross profit, though the framework supports any financial KPI, such as revenue, quick ratio, or cash flow). The objective is to produce a single, structured recommendation $a$ that, if implemented, would improve $g$. Each recommendation is emitted as a JSON object with a \texttt{recommendation} (the concrete action), a \texttt{reasoning} field, and a quantified \texttt{expected\_impact}.

We treat advice generation as a reinforcement learning problem and learn a policy $\pi_\theta(a \mid s, g)$ that maps a state and goal to a recommendation. This framing is motivated by two properties of the domain. First, the historical actions logged for each business are not optimal targets, as they reflect the constraints and incomplete information available at decision time, so supervised imitation would propagate suboptimal behavior. Second, recommendations are open-ended natural language, ruling out value-based methods over a fixed action set. RL instead lets us optimize a scalar reward that directly encodes what makes a recommendation useful.

\subsection{GRPO Training}
\label{sec:grpo}

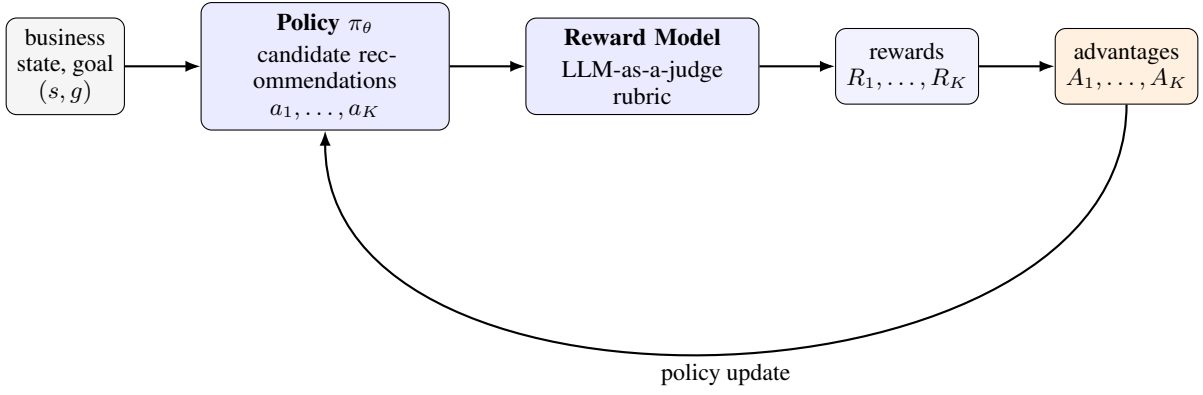
\begin{figure*}[t]
  \centering
  \begin{tikzpicture}[
    font=\small,
    box/.style={draw, rounded corners, align=center, inner sep=4pt, minimum height=10mm},
    pol/.style={box, fill=blue!8, text width=30mm},
    rew/.style={box, fill=blue!8, text width=28mm},
    grp/.style={box, fill=blue!5, minimum width=12mm},
    adv/.style={box, fill=orange!12, minimum width=12mm},
    arr/.style={-{Latex[length=2mm]}, thick},
  ]
    \node[box, fill=gray!8] (state) {business\\ state, goal\\ $(s, g)$};
    \node[pol, right=10mm of state] (policy)
      {\textbf{Policy} $\pi_\theta$\\[2pt]\footnotesize candidate recommendations\\$a_1, \dots, a_K$};
    \node[rew, right=10mm of policy] (reward)
      {\textbf{Reward Model}\\[2pt]\footnotesize LLM-as-a-judge\\rubric};
    \node[grp, right=10mm of reward] (rewards) {rewards\\$R_1, \dots, R_K$};
    \node[adv, right=10mm of rewards] (adv) {advantages\\$A_1, \dots, A_K$};

    \draw[arr] (state) -- (policy);
    \draw[arr] (policy) -- (reward);
    \draw[arr] (reward) -- (rewards);
    \draw[arr] (rewards) -- (adv);
    \draw[arr] (adv) to[out=-90, in=-90] node[below, pos=0.5] {\footnotesize policy update} (policy);
  \end{tikzpicture}
  \caption{GRPO training loop. For each business state and goal $(s, g)$, the policy samples $K$ candidate recommendations, each is scored by the LLM-as-a-judge rubric reward, and group-relative advantages within the group drive the policy update.}
  \label{fig:grpo}
\end{figure*}

We optimize $\pi_\theta$ with Group Relative Policy Optimization \citep{shao2024deepseekmath}. For each prompt $(s, g)$ we sample a group of $K$ candidate recommendations $\{a_1, \dots, a_K\}$ from the current policy, score each with the reward function of Section~\ref{sec:reward} to obtain rewards $\{R_1, \dots, R_K\}$, and compute a group-relative advantage by standardizing rewards within the group,
\begin{equation}
A_i = \frac{R_i - \operatorname{mean}(R_1, \dots, R_K)}{\operatorname{std}(R_1, \dots, R_K)}.
\end{equation}
The policy is then updated with the clipped GRPO objective, which avoids training a separate value model. Figure~\ref{fig:grpo} illustrates this loop. We additionally anchor the policy to the base model with a KL penalty weighted by $\beta$, which we found necessary to prevent policy drift during training (Section~\ref{sec:reward}).

\subsection{Reward Design}
\label{sec:reward}

The reward is provided by an LLM judge (Claude Opus 4.5) that scores a recommendation against a rubric of eleven binary criteria: a safety gate plus ten quality criteria organized into six aspects, summarized in Table~\ref{tab:rubric}. Each criterion is judged independently as $0$ or $1$.

\begin{table}[t]
  \centering
  \small
  \begin{tabular}{@{}p{15mm}c p{40mm}@{}}
    \toprule
    \textbf{Aspect} & \textbf{\#} & \textbf{Binary criterion} \\
    \midrule
    Safety (gate) & 1 & Would not cause significant harm to the business \\
    \addlinespace
    Specificity & 2 & Cites a concrete number from the state \\
                & 3 & Names a specific entity (vendor, account, line item) \\
    \addlinespace
    Actionability & 4 & Implementable immediately without further analysis \\
    \addlinespace
    Data grounding & 5 & All stated facts are quoted from or derived from the state \\
    \addlinespace
    Reasoning & 6 & Identifies a non-obvious pattern or anomaly \\
              & 7 & Acknowledges the primary risk of the action \\
    \addlinespace
    Impact & 8 & Expected impact is quantified \\
           & 9 & Impact estimate is grounded in a state value \\
           & 10 & Impact is directionally correct \\
    \addlinespace
    Relevance & 11 & Targets an item that can materially move the goal metric \\
    \bottomrule
  \end{tabular}
  \caption{The eleven binary rubric criteria: a safety gate (criterion~1) plus ten quality criteria (criteria~2--11) across six aspects. The relevance criterion is goal-conditioned, adapting to the active target metric.}
  \label{tab:rubric}
\end{table}

The safety gate is critical: if implementing the recommendation would cause significant harm to the business (e.g., eliminating a primary revenue stream or creating unsustainable cash-flow risk), the gate hard-zeros the reward,
\begin{equation}
\label{eq:critic}
c(a) =
\begin{cases}
0 & \text{if unsafe}, \\[2pt]
\dfrac{1}{D} \sum_{d=1}^{D} \mathbf{1}[d \text{ satisfied}] & \text{otherwise},
\end{cases}
\end{equation}
where $D = 10$ is the number of quality criteria. The rubric is goal-conditioned: the relevance criterion adapts to the active target metric.

Two penalties shape the final reward. Recommendations whose JSON cannot be parsed receive a fixed penalty of $-0.5$. In addition, to discourage degenerate over-long reasoning, a thinking-length penalty $p(a) \in [0, 0.2]$ is applied that is zero below a soft cap on reasoning length and ramps linearly to its maximum at a hard cap. The final reward is
\begin{equation}
R(a) = c(a) - p(a).
\end{equation}
Both terms address failure modes we observed in early runs: without the KL anchor the policy drifted from the base model, and without the length penalty reasoning traces grew unboundedly until JSON outputs collapsed.

\subsection{Judge-Independent Causal Audit}
\label{sec:cate-method}

The training reward comes from an LLM judge. As a result, a higher judge score may reflect recommendations that appeal to the judge rather than genuinely better business outcomes. To test for this, we audit the learned policy with a judge-independent estimate of its effect on real outcomes. This estimate is computed entirely from logged observational data.

The key step is to bridge from free-text recommendations to a quantity for which causal effects can be estimated. An independent action mapper (Section~\ref{sec:cate-impl}) assigns each recommendation to a discrete business action; the resulting action is treated as the \emph{treatment}, the business-state embedding as the conditioning covariates, and the realized year-over-year gross-profit growth as the outcome. We then estimate the conditional average treatment effect (CATE) of the policy's actions using a standard doubly-robust (AIPW) estimator \citep{robins1994estimation, bang2005doubly, dudik2011doubly, abrevaya2015estimating}, and report the estimated lift alongside a downside rate and a tail-risk measure. We use these estimates as a judge-independent policy-ranking audit rather than as evidence of realized, deployed business impact. The estimator, its propensity and outcome models, and the action mapper are detailed in Section~\ref{sec:experiments}.

\section{Experiments}
\label{sec:experiments}

\subsection{Experimental Setup}
\label{sec:setup}

\paragraph{Data and model.} We train on logged business financial states paired with the goal of improving gross profit. The base policy is Qwen3.5-27B, fine-tuned with LoRA under DeepSpeed ZeRO-2. Personally identifying entities are removed from the training data and replaced with synthetic surrogates: rather than leaving opaque mask placeholders in the state, the pipeline substitutes realistic synthetic vendor, account, and product names, so that the model is trained on well-formed business text while no real customer entity is retained. GRPO uses $K{=}12$ candidate generations per prompt, a maximum completion length of $8{,}000$ tokens, a KL coefficient $\beta{=}0.001$, learning rate $5\!\times\!10^{-5}$ with a cosine schedule, and the doubly-robust GRPO loss variant. The judge reward model is Claude Opus 4.5. We select the best checkpoint on a held-out validation reward.

\paragraph{Evaluation protocols.} We use two complementary evaluations. The first is the LLM-as-a-judge rubric score (Section~\ref{sec:reward}), reported on a held-out set of $500$ businesses with $5$ independent trials per business ($n{=}2{,}500$ per model); we report the mean over per-trial means with a $95\%$ confidence interval. The second is the causal off-policy evaluation (Section~\ref{sec:cate-method}), which we instantiate below. We compare against commercial LLMs (Claude Opus 4.5/4.6, Claude Sonnet 4.5, GPT-5.4) and the untrained Qwen3.5-27B base model.

\subsection{Causal Estimator Instantiation}
\label{sec:cate-impl}

We estimate the effect of a policy's recommended action on year-over-year gross-profit growth from logged data. Treatment is defined \emph{per action}: for a target action $a$, let $A \in \{0,1\}$ indicate whether $a$ was taken, so that the treated arm consists of logged states in which $a$ was taken and the control arm consists of states in which a \emph{different} catalogue action was taken. A single multi-label propensity model supplies $e(x) = P(A{=}1 \mid x)$ for every action, and each action is stratified independently. For a business with covariates $x$ (a $768$-dimensional embedding of the business state, encoded with a sentence-embedding model), let $Y$ denote the realized year-over-year gross-profit growth. We fit a propensity model $\hat{e}$ (a multi-label MLP over the state embedding) and an outcome model with separate treated and control heads $\hat{\mu}_1, \hat{\mu}_0$ (neural networks predicting $Y$ under the action and under its absence). The doubly-robust AIPW pseudo-outcome for each logged business is
\begin{equation}
\label{eq:aipw}
\begin{split}
\phi_i = \big(\hat{\mu}_1(x_i) - \hat{\mu}_0(x_i)\big)
 &+ \frac{A_i}{\hat{e}(x_i)}\big(Y_i - \hat{\mu}_1(x_i)\big) \\
 &- \frac{1 - A_i}{1 - \hat{e}(x_i)}\big(Y_i - \hat{\mu}_0(x_i)\big),
\end{split}
\end{equation}
which is unbiased if \emph{either} the propensity or the outcome model is correctly specified \citep{bang2005doubly}.

\paragraph{What is computed per recommendation.} The audit proceeds in two stages, and it is worth stating precisely which quantity each stage produces. In \emph{stage one}, on logged data where the treatment $A_i$ and the outcome $Y_i$ are both observed, we evaluate the AIPW pseudo-outcome of Eq.~\ref{eq:aipw} for every logged business and average it within each (action, propensity-decile) stratum. Strata that fail minimum-support or effective-sample-size gates are marked invalid and excluded. This yields a table of stratum-level effect estimates spanning the $49$ actions that clear the eligibility threshold and $10$ propensity bins per action. In \emph{stage two}, each held-out recommendation is mapped to an action, the propensity of its business state for that action is predicted and assigned to a bin, and the recommendation receives the corresponding stratum-level effect. Equation~\ref{eq:aipw} is therefore evaluated only on logged businesses; a held-out recommendation receives a stratum-level lookup rather than its own pseudo-outcome, because advice that was never acted upon has no observed $A_i$ or $Y_i$. The estimand is thus the mean stratum-level effect of the (action, propensity-bin) cells into which a policy's recommendations fall, and policies differ only through which actions they select and how their states distribute across propensity bins.

The three reported metrics are computed over this set of per-recommendation values. The estimated lift (DR-CATE) is their mean. The \emph{downside rate} (DR\%) is the fraction whose assigned stratum effect is negative. The conditional value-at-risk $\mathrm{CVaR}_{0.10}$ is the mean over the worst $10\%$. Higher lift is better; a lower downside rate and a less negative $\mathrm{CVaR}_{0.10}$ are better.

The audit follows the same protocol as the judge evaluation. Each policy is evaluated on $500$ randomly sampled held-out business states over $5$ independent runs with different random seeds. We compute the three metrics within each run, and report the mean across runs together with a $95\%$ confidence interval over the five run-level estimates. Reporting uncertainty at the level of runs rather than individual recommendations means the intervals reflect run-to-run variation in the policy's own sampling, which is the variation that matters when comparing policies. The mapper, the propensity model, and the outcome model are held fixed across all policies, so every system is scored by the same instruments.

\paragraph{Action mapper.} Recommendations are mapped to the action taxonomy by a separate LLM classifier, distinct from and independent of the Claude Opus 4.5 reward judge, so that mapping noise cannot directly inflate the audit. Given a recommendation and the fixed action catalogue, the mapper returns the single best-matching action together with a confidence score; recommendations that fall below a confidence threshold, or that match no catalogue action, are assigned a dedicated \emph{no-match} label and excluded from scoring rather than forced onto an ill-fitting action. We summarize the mapper here and will release a synthetic, anonymized version with the reproducibility artifacts.

A compact summary of the evaluation setup is given in Table~\ref{tab:datacard}.

\begin{table}[t]
  \centering
  \small
  \begin{tabular}{@{}ll@{}}
    \toprule
    \textbf{Item} & \textbf{Value} \\
    \midrule
    Evaluation unit      & company-month \\
    Sample per run       & 500 business states \\
    Runs per policy      & 5 (different seeds) \\
    Action catalogue     & 60 actions \\
    Action categories    & 10 \\
    Outcome              & YoY gross-profit growth \\
    Causal estimator     & doubly-robust AIPW (DR-CATE) \\
    Propensity model     & multi-label MLP \\
    Outcome model        & neural, treated/control heads \\
    State embedding      & 768-dim sentence embedding \\
    Holdout split        & company-level \\
    Free-text mapping    & LLM mapper (no-match option) \\
    \bottomrule
  \end{tabular}
  \caption{Evaluation/data card for the judge-independent causal audit.}
  \label{tab:datacard}
\end{table}

\subsection{Results}
\label{sec:results}

\paragraph{Judge evaluation.} Table~\ref{tab:judge} reports the rubric scores. After GRPO training, Qwen3.5-27B reaches $9.514$, the top score on the leaderboard, ahead of Claude Opus 4.6 ($9.365$), GPT-5.4 ($8.949$), and Claude Sonnet 4.5 ($8.712$), and well above the untrained base model ($8.457$). Two caveats apply to this table. First, the judge belongs to the Claude family, which may introduce a small self-preference bias in favor of Claude baselines; this makes the comparison conservative for our model. Second, and more importantly, our policy was trained against this same judge, so its rubric score is in part a measure of successful optimization rather than an independent assessment. We therefore read Table~\ref{tab:judge} as evidence that training achieved its objective, and treat the judge-independent causal audit below as the primary comparative evidence.

\begin{table}[t]
  \centering
  \small
  \begin{tabular}{@{}lcc@{}}
    \toprule
    \textbf{Model} & \textbf{Score} & \textbf{95\% CI} \\
    \midrule
    \textbf{Qwen3.5-27B-GRPO (ours)} & \textbf{9.514} & [9.505, 9.524] \\
    Claude Opus 4.6 & 9.365 & [9.354, 9.376] \\
    Claude Opus 4.5 & 8.982 & [8.954, 9.010] \\
    GPT-5.4 & 8.949 & [8.911, 8.988] \\
    Claude Sonnet 4.5 & 8.712 & [8.644, 8.780] \\
    Qwen3.5-27B (base) & 8.457 & [8.384, 8.531] \\
    \bottomrule
  \end{tabular}
  \caption{LLM-as-a-judge rubric scores, $n{=}2{,}500$ per model. Scores are the mean number of the $D{=}10$ quality criteria satisfied, i.e.\ the quality term $c(a)$ of Eq.~\ref{eq:critic} rescaled to $[0, 10]$; the safety gate zeroes the score of any recommendation judged unsafe. Our GRPO-trained model ranks first.}
  \label{tab:judge}
\end{table}

\paragraph{Causal evaluation.} Table~\ref{tab:cate} reports the causal off-policy results. Our policy attains an estimated gross-profit lift of $0.0228$ $[0.0211, 0.0246]$, approximately twice that of the strongest commercial baseline, Claude Opus 4.6 ($0.0104$, a ratio of $2.20\times$), together with the lowest downside rate ($0.155$ vs.\ $0.232$) and the least negative tail risk ($\mathrm{CVaR}_{0.10} = -0.073$ vs.\ $-0.100$), meaning that even its worst-decile recommendations are less harmful on average. Our policy ranks first on all three metrics, and its confidence interval does not overlap that of any other policy on any metric; under a Welch $t$-test over the five per-trial means, every pairwise comparison with our policy is significant at $p < 0.01$.

The commercial baselines separate clearly under the audit. Claude Opus 4.6 is the only commercial system with a lift that is both positive and comfortably distinguishable from zero ($p < 0.001$); GPT-5.4 is positive but noisier ($0.0082$, $p = 0.015$, and the widest interval of any policy); Claude Sonnet 4.5 is not distinguishable from zero ($0.0028$ $[0.0000, 0.0055]$, $p = 0.051$); and Claude Opus 4.5 is estimated to be \emph{negative} ($-0.0025$ $[-0.0043, -0.0007]$), i.e.\ its recommendations map to actions that are, on average, associated with slightly worse subsequent gross-profit growth. Downside rate tracks this ordering closely: the three weakest policies place roughly a third of their recommendations on actions with negative estimated effect, against $15.5\%$ for ours.

\begin{table*}[t]
  \centering
  \small
  \setlength{\tabcolsep}{6pt}
  \begin{tabular}{@{}lccc@{}}
    \toprule
    \textbf{Policy} & \textbf{Lift (DR-CATE)} $\uparrow$ & \textbf{DR\%} $\downarrow$ & $\mathbf{CVaR_{0.10}}$ $\uparrow$ \\
    \midrule
    \textbf{Qwen3.5-27B-GRPO (ours)} & $\mathbf{0.0228 \pm 0.0017}$ & $\mathbf{0.155 \pm 0.017}$ & $\mathbf{-0.073 \pm 0.003}$ \\
    Qwen3.5-27B (base)   & $0.0170 \pm 0.0024$ & $0.194 \pm 0.018$ & $-0.094 \pm 0.009$ \\
    \addlinespace
    Claude Opus 4.6      & $0.0104 \pm 0.0031$ & $0.232 \pm 0.021$ & $-0.100 \pm 0.008$ \\
    GPT-5.4              & $0.0082 \pm 0.0056$ & $0.327 \pm 0.040$ & $-0.124 \pm 0.007$ \\
    Claude Sonnet 4.5    & $0.0028 \pm 0.0028$ & $0.320 \pm 0.027$ & $-0.114 \pm 0.006$ \\
    Claude Opus 4.5      & $-0.0025 \pm 0.0018$ & $0.362 \pm 0.021$ & $-0.116 \pm 0.006$ \\
    \bottomrule
  \end{tabular}
  \caption{Judge-independent causal off-policy evaluation: estimated gross-profit lift (DR-CATE), downside rate (DR\%), and tail risk ($\mathrm{CVaR}_{0.10}$). Policies are ordered by lift. Each entry is the mean over $5$ independent runs ($500$ sampled business states per run, different seed per run) $\pm$ the half-width of a $95\%$ confidence interval computed over the five run-level estimates. Higher lift and higher (less negative) $\mathrm{CVaR}_{0.10}$ are better; lower DR\% is better. Our GRPO policy ranks first on all three metrics, with no confidence-interval overlap against any other policy on any metric.}
  \label{tab:cate}
\end{table*}

\paragraph{Effect of GRPO training.} Comparing our policy against its own starting point isolates the contribution of GRPO from that of the base model. Training raises the estimated lift from $0.0170$ to $0.0228$, a relative improvement of $34\%$ (Welch $t$-test over the five run-level estimates, $p = 0.0009$; because all policies are evaluated under the same five seeds, a paired test on the same estimates gives $p = 0.003$), while cutting the downside rate from $0.194$ to $0.155$ ($p = 0.002$) and improving tail risk from $-0.094$ to $-0.073$ ($p = 0.002$). The gain is therefore not merely a property of the base checkpoint: GRPO with the safety-gated rubric reward moves the policy toward actions with higher estimated effect \emph{and} away from actions with negative estimated effect, which is precisely the behavior the safety gate was designed to induce.

\paragraph{The two evaluations do not rank identically.} The audit and the judge agree that our GRPO policy is best, but they disagree elsewhere, and the disagreement is informative. The untrained Qwen3.5-27B base model ranks \emph{last} on the judge rubric ($8.457$, Table~\ref{tab:judge}) yet \emph{second} on the causal audit ($0.0170$, Table~\ref{tab:cate}), above every commercial system. Conversely, Claude Opus 4.5 scores well on the rubric ($8.982$) but has a negative estimated lift. We read this as evidence that the two evaluations measure genuinely different things. The rubric rewards well-formed advice: concrete figures, named entities, quantified impact, acknowledged risk. The audit rewards advice that maps to actions historically associated with gross-profit growth, and is indifferent to how well that advice is written. Commercial models are strong at the former and only middling at the latter, whereas the base model produces less polished recommendations that nonetheless land on reasonable actions. This divergence is the clearest available evidence that the audit is not a proxy for the judge; had it simply re-scored rubric quality, the base model would have ranked last on both. It also means the joint result carries more weight than either evaluation alone: our policy is the only system that ranks first under both.

\section{Discussion}
Three findings stand out. First, GRPO with a finance-specific reward and explicit safety gate lifts an open-weight model above strong commercial LLMs on the judge rubric (Table~\ref{tab:judge}). This indicates that targeted reward design can substitute for raw scale on a narrow, high-value task. Second, the causal audit, which is independent of the judge, moves in the same direction (Table~\ref{tab:cate}): the GRPO policy attains a higher estimated gross-profit lift, a lower downside rate, and a less negative tail risk than every commercial baseline, with non-overlapping confidence intervals throughout. Because the audit is computed from logged outcomes and does not rely on the reward judge, this agreement makes pure judge adaptation a less likely explanation for the gains, although it does not establish realized impact.

Third, and less expected, the two evaluations rank the \emph{other} systems quite differently. That the untrained base model is last on the rubric but second on the audit shows the audit is not a restatement of the judge, and it also cautions against reading either evaluation as a complete measure of advice quality on its own: a policy can produce recommendations that are well-formed but point at unhelpful actions, or poorly-formed but point at sensible ones. Only a system that scores well on both, as ours does, is supported by both lines of evidence.

The safety gate and the risk-aware metrics are complementary. The gate suppresses harmful recommendations during training, while the downside rate and tail-risk metrics verify after training that the resulting policy places less mass on actions with negative estimated effects; the drop in downside rate from $0.194$ (base) to $0.155$ (ours) is direct evidence that this mechanism operates as intended.

\section{Conclusions}
We formulate financial advice generation as a reinforcement learning problem and train an open-weight model with GRPO using a finance-specific reward that incorporates explicit safety constraints. The resulting model outperforms the strongest evaluated commercial baseline on both the judge rubric and the CATE audit, and is the only system to rank first under both. It achieves approximately twice the estimated gross-profit lift while maintaining the lowest downside rate and least negative tail risk of any policy evaluated, and improves over its own starting checkpoint by $34\%$ in estimated lift. We also find that the judge and the causal audit rank the baselines differently, which suggests that rubric quality and estimated business value are distinct axes and that reporting only one of them can be misleading. These results suggest that careful reward design, combined with a judge-independent causal audit to reduce the risk of reward hacking, provides a practical approach for building safer and more effective financial advice models. Future work will explore richer action taxonomies, evaluation by financial experts, and support for multi-step advice.

\section*{Limitations}
Because randomized experiments are not available in this setting, our causal audit is computed from logged observational data, in common with off-policy evaluation more broadly. We therefore treat it as a judge-independent, complementary signal alongside the rubric score rather than as a standalone measure, and the two evaluations are designed to be read together. A second limitation is coverage. Recommendations that receive no confident match in the $60$-action catalogue are excluded from the audit rather than forced onto an ill-fitting action. This keeps the mapping conservative, but it means each policy is audited on the subset of its own output that the catalogue can express, and a policy whose unmatched recommendations differ systematically in quality from its matched ones would be scored on a non-representative subset. A richer catalogue would narrow this gap. Our audit also scores a single recommended action per business state; extending it to multi-step advice, where actions compound over a longer horizon, is a natural direction for future work. Finally, our study targets gross profit as the primary goal metric; while the framework supports other financial KPIs, we leave a broad multi-KPI evaluation to future work.


\bibliography{custom}

\appendix


\end{document}